\documentclass[sigconf, authorversion]{aamas} 
\usepackage{balance} 
\usepackage{siunitx}
\usepackage{booktabs}
\usepackage{subcaption}

\title{Curiosity-Driven Multi-Agent Exploration\\ with Mixed Objectives}

\author{Roben Delos Reyes, Kyunghwan Son, Jinhwan Jung, Wan Ju Kang, Yung Yi}
\affiliation{
  \department{School of Electrical Engineering}
  \institution{Korea Advanced Institute of Science and Technology}}
\email{{rddelosreyes, kevinson9473, mp3jjk, soarhigh, yiyung}@kaist.ac.kr}


\newcommand{\BibTeX}{\rm B\kern-.05em{\sc i\kern-.025em b}\kern-.08em\TeX}

\begin{document}


\pagestyle{fancy}
\fancyhead{}

\newcommand{\algname}{{\text{TWO}}}

\begin{abstract}
Intrinsic rewards have been increasingly used to mitigate the sparse reward problem in single-agent reinforcement learning. These intrinsic rewards encourage the agent to look for novel experiences, guiding the agent to explore the environment sufficiently despite the lack of extrinsic rewards. Curiosity-driven exploration is a simple yet efficient approach that quantifies this novelty as the prediction error of the agent's curiosity module, an internal neural network that is trained to predict the agent's next state given its current state and action. We show here, however, that na\"ively using this curiosity-driven approach to guide exploration in sparse reward cooperative multi-agent environments does not consistently lead to improved results. Straightforward multi-agent extensions of curiosity-driven exploration take into consideration either individual or collective novelty only and thus, they do not provide a distinct but collaborative intrinsic reward signal that is essential for learning in cooperative multi-agent tasks. In this work, we propose a curiosity-driven multi-agent exploration method that has the mixed objective of motivating the agents to explore the environment in ways that are individually and collectively novel. First, we develop a two-headed curiosity module that is trained to predict the corresponding agent's next observation in the first head and the next joint observation in the second head. Second, we design the intrinsic reward formula to be the sum of the individual and joint prediction errors of this curiosity module. We empirically show that the combination of our curiosity module architecture and intrinsic reward formulation guides multi-agent exploration more efficiently than baseline approaches, thereby providing the best performance boost to MARL algorithms in cooperative navigation environments with sparse rewards.
\end{abstract}

\maketitle 

\section{Introduction}
\begin{figure*}[t!]
  \centering
  \includegraphics[scale=.45]{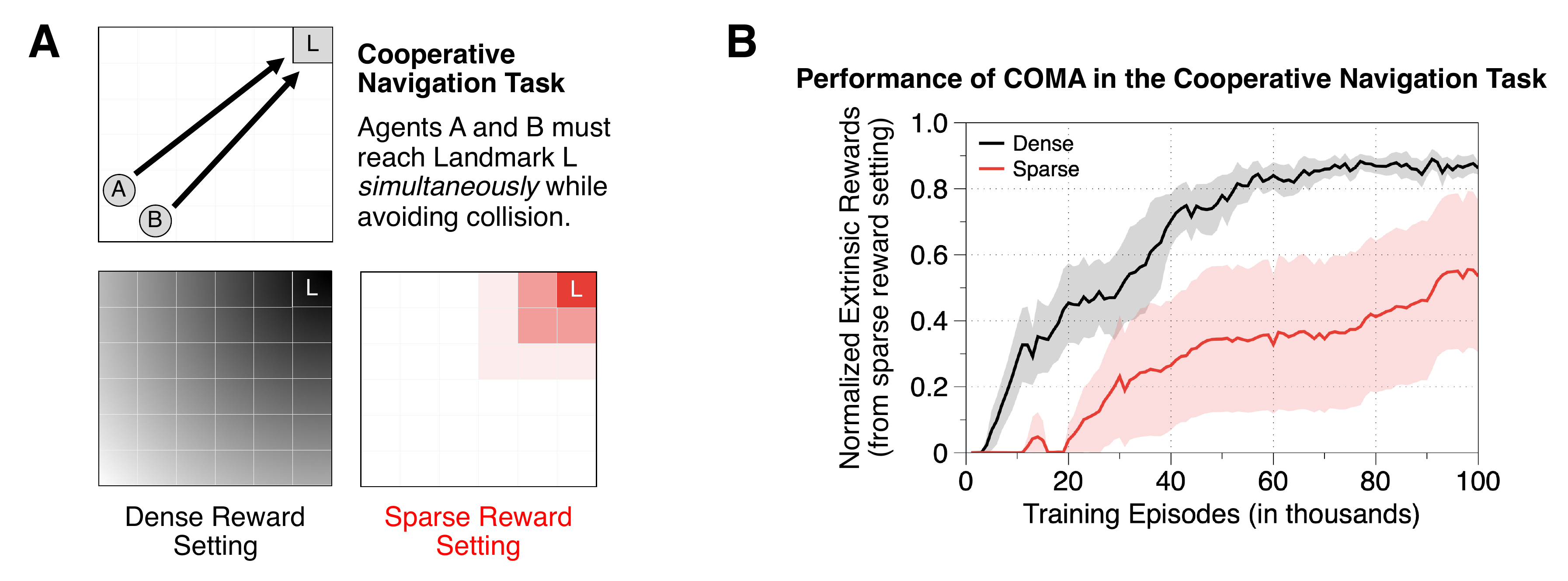}
  \caption{(A) Consider the cooperative navigation task. In the dense reward setting, the agents receive a distinct extrinsic reward anywhere in the environment, but in the sparse reward setting, the agents only receive an extrinsic reward when they are both near the landmark. (B) COMA \cite{coma} agents trained in a dense reward environment learn to solve the task more consistently than COMA agents trained in a sparse reward environment.}
  \label{fig:fig1}
  \Description{Consider the cooperative navigation task. In the dense reward setting, the agents receive a distinct reward anywhere in the environment, but in the sparse reward setting, they only receive a reward when they are both near the landmark. (B) Across 10 seeds, COMA \cite{coma} agents trained in a dense reward environment consistently learn how to reach the landmark after about 20,000 training episodes, whereas COMA agents trained in a sparse reward environment find it difficult to solve the task even when trained five times longer.}
\end{figure*}

Deep multi-agent reinforcement learning (MARL) has emerged as a standard framework for training autonomous agents to accomplish cooperative tasks. State-of-the-art results were reported by cooperative MARL algorithms \cite{maddpg,maac,coma,iql,vdn,qmix,qtran} in various multi-agent environments like in cooperative navigation scenarios \cite{particle_environment} and in real-time strategy games \cite{smac}. However, these results were observed in dense reward settings where agents continuously receive well-designed reward signals from the environment, such as the distance between agents and their destinations. As shown in Figure \ref{fig:fig1}, the performance of these algorithms degrades significantly in the sparse reward setting where rewards are only received by the agent at hard-to-reach areas of the environment or during late stages of the task. The lack of distinct feedback from the environment on how agents should behave early on the learning process exacerbates the difficulty of learning optimal cooperative policies. 

The difficulty of learning in sparse reward environments is associated with a problem with exploration. Sparse reward environments are desirable because designing a good reward function for a task is not always straightforward. But because agents infrequently receive feedback on how to behave accordingly, agents lack the motivation and direction on how to explore the environment. In the single-agent reinforcement learning (RL) framework, the agent is trained to additionally maximize intrinsic rewards when extrinsic rewards are sparse. These intrinsic rewards are generated by the agent itself based on its knowledge of the environment and are formulated such that the agent receives high intrinsic rewards for having novel experiences, e.g., visiting states with low visitation counts \cite{hashtag_exploration,unifying,neural_density_models} or going to areas of the environment where the state transition dynamics are unfamiliar to the agent \cite{icm,large-scale,disagreement,rnd,ride}. When rewards from the environment are sparse, intrinsic rewards serve as effective exploration bonuses that guide the agent in exploring the environment.

Curiosity-driven exploration is a simple yet efficient approach which uses intrinsic rewards to encourage exploration in single-agent environments with sparse rewards. The agent is equipped with an internal neural network that is trained to predict the agent's next state $s_{t+1}$ given its current state $s_t$ and action $u_t$. This neural network, also known as a curiosity module, represents the agent's knowledge of the environment. Thus, it outputs a high prediction error for regions of the environment whose transition dynamics are not yet well-known to the agent. Curiosity-driven exploration methods view this prediction error as a form of curiosity and provide it to the agent as an intrinsic reward. Existing works demonstrated that curiosity as an intrinsic reward significantly improves the performance of single-agent RL algorithms in various hard exploration tasks \cite{icm,large-scale,disagreement,rnd,ride,lwm}.

However, we show in this work that na\"ively using curiosity-driven exploration for sparse reward multi-agent environments is not as effective as it is for single-agent scenarios. There are two straightforward ways to utilize curiosity-driven exploration methods for MARL. First, we can equip each agent with its own curiosity module and use the individual prediction error of this module as the agent's intrinsic reward \cite{coordinated_exploration, synergy}. This approach rewards each agent for having individually novel experiences which motivates the agents to explore the environment regardless of what other agents already know about it. For tasks that require cooperation among many agents, this independent exploration strategy is not necessarily suitable. On the other hand, we can design the intrinsic reward formula to quantify collective novelty instead, e.g., by using a joint prediction error \cite{synergy} or by taking the minimum of individual prediction errors \cite{coordinated_exploration}. Being a collective measure, this intrinsic reward signal does induce cooperation naturally, just like why cooperative multi-agent environments often provide a shared team reward to the agents \cite{particle_environment, smac}. However, using a collective reward suffers from the credit assignment problem wherein agents find it difficult to assess from the team reward how well each of them behaved.

In this work, we address these shortcomings by proposing a curiosity-driven multi-agent exploration method that has the mixed objective of motivating the agents to explore the environment in ways that are individually and collectively novel. In particular, we provide the following new contributions:
\begin{itemize}
    \item We develop a two-headed curiosity module that predicts the corresponding agent's next observation in the first head and the next joint observation in the second head. 
    \item We design the intrinsic reward formula to be the sum of the individual prediction error and the joint prediction error of the agent's two-headed curiosity module. 
    \item We show that the combination of our curiosity module architecture and intrinsic reward formulation best improves the performance of MARL algorithms in cooperative navigation environments with sparse rewards.
\end{itemize}

\section{Related Work}
The goal of an RL agent is to maximize the rewards that it receives from the environment. It is these rewards as well that help the agent learn how to solve the given task. When these extrinsic rewards are sparse, the agent will naturally lose the motivation and direction on how to explore the environment. In such cases, providing the agent with intrinsic rewards has been shown to be an effective way to encourage exploration.

\subsection{Encouraging Single-Agent Exploration\\ with Intrinsic Rewards}
Intrinsic rewards are seen as a form of intrinsic motivation \cite{oudeyer_intrinsic}. They are able to encourage an RL agent to explore the environment even when it barely receives any useful learning feedback. For these scenarios, the agent is trained to maximize instead the sum of the sparse extrinsic reward $e_t$ and the dense intrinsic reward $i_t$:
\begin{equation*}
r_t = e_t + \lambda i_t,
\end{equation*}
where the trade-off coefficient $\lambda$ weighs the importance of each reward. The intrinsic reward $i_t$ is generated by the agent itself based on its knowledge of the environment and is formulated in a way that it has a high value for experiences that are novel to the agent. 

For example, a classic exploration method for single-agent RL defines the intrinsic reward $i_t$ as the inverse square root of the state-action visitation count, which is stored in a table, in order to motivate the agent to visit less frequently visited states \cite{mbie-eb}. This intrinsic reward formulation has been shown to be effective for environments with small, finite state spaces, and it even has theoretical guarantees. However, as RL started to incorporate the use of deep neural networks, it also began to use larger and more complex environments where an agent rarely visits the same state a number of times. For these large environments, maintaining a table of visitation counts is no longer practical.
 
Since then, many works have proposed new ways of generating and measuring these intrinsic rewards so that they can still be effective for environments with a large state space. A set of these works known as count-based methods extends the classic visitation counts to handle large environments by using hash tables \cite{hashtag_exploration} or density models \cite{unifying, neural_density_models}. Other papers motivate the agent to maximize information gain instead \cite{vime, emi, diversity}. There is also a class of methods which uses the prediction error of a neural network as an intrinsic reward \cite{icm,rnd,disagreement,ride,lwm, large-scale}. Regardless of the approach, these exploration methods significantly improve the performance of single-agent RL algorithms in various hard exploration tasks.  

\subsection{Cooperative Multi-Agent Exploration}
Although exploration in single-agent RL has fairly been studied over the years, multi-agent exploration has just recently gained the attention of the MARL community. Unlike its single-agent counterpart, exploration in MARL faces unique challenges due to the nature of cooperative multi-agent systems. Aside from having to deal with the exponential increase in the size of the state and action spaces, the works in this research area are primarily concerned with identifying a mechanism that can effectively induce cooperative exploration. This is because solving cooperative multi-agent tasks requires cooperation among many agents, and in order to efficiently manage that, agents must also learn to coordinate how they explore the environment.

Some works attempted to mitigate this problem by having a training scheme that utilizes a representation that is shared across the agents, such as by using a shared latent variable \cite{maven} or by having shared goals \cite{cmae}. Quite notably, majority of these works still make use of intrinsic rewards which were adapted from single-agent exploration methods. The works of B\"{o}hmer et al. \cite{unreliable} and Iqbal and Sha \cite{coordinated_exploration} use count-based methods to make learning with intrinsic rewards reliable and adaptive in MARL, respectively. On the other hand, Wang et al. \cite{edti} and Jiang and Lu \cite{eoi} show that information gain can still be an effective intrinsic reward signal in the multi-agent case. There is also the work of Chitnis et al. \cite{synergy} which applies the curiosity-driven approach to motivate the agents to perform synergistic behaviors.

Among these works, it is the work of Chitnis et al. \cite{synergy} that is closest to our proposed method. Similar to them, we use prediction error as an intrinsic reward to drive exploration in cooperative MARL. However, the method of Chitnis et al. is only trainable under the fully centralized paradigm. For many multi-agent scenarios, using the fully centralized paradigm to train multiple agents is not optimal because it is not scalable and it makes strong assumptions on the amount of information that is readily available to the agents when they are choosing their actions. What is more commonly used in many MARL papers is the paradigm of centralized training with decentralized execution (CTDE) \cite{coma,maddpg}. So in this paper, we propose a curiosity-driven multi-agent exploration method that is trained under the CTDE paradigm. 
\section{Background}

\begin{figure}[b!]
  \centering
  \includegraphics[scale=0.45]{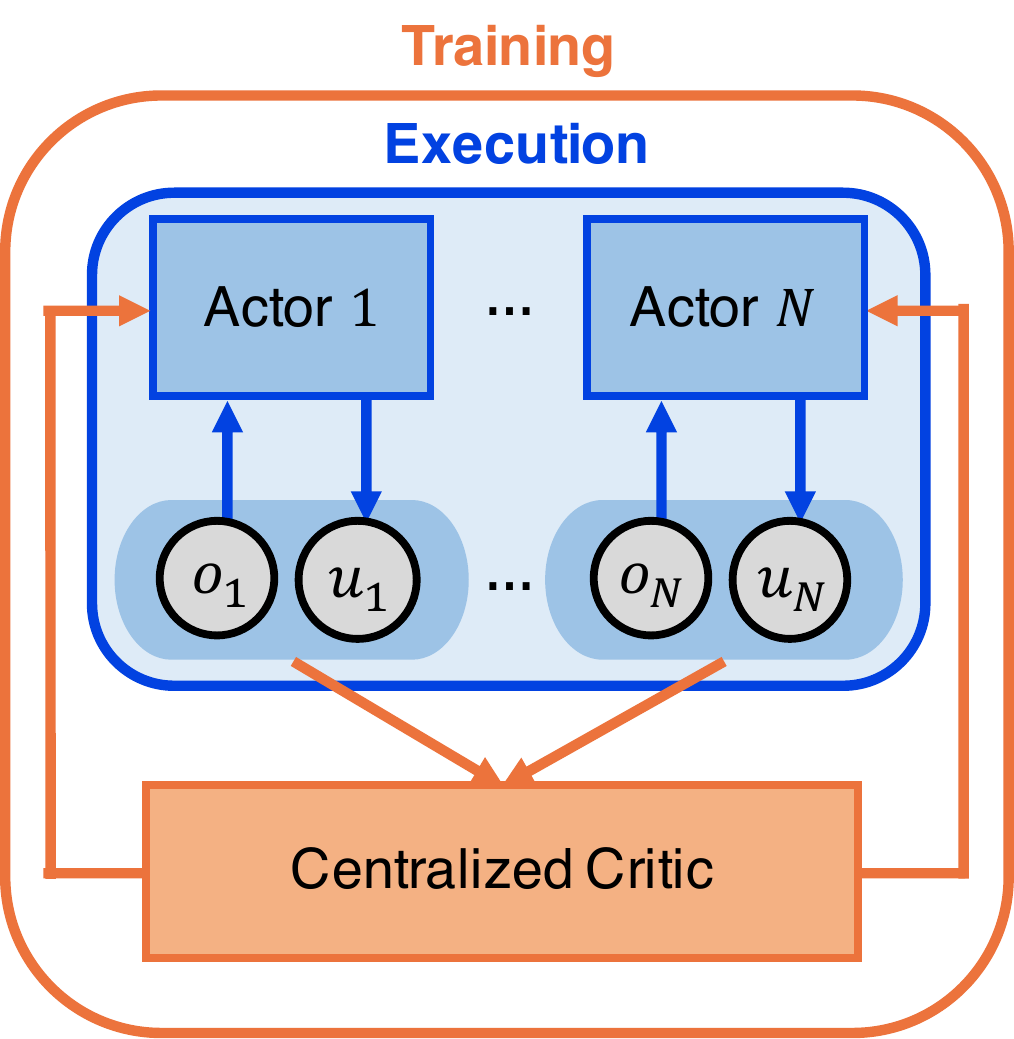}
  \caption{Overview of the CTDE paradigm. Many works on MARL use the CTDE paradigm because it offers a stable and scalable framework for training many agents.}
  \label{fig:fig2}
  \Description{Overview of the CTDE paradigm. Many works on MARL are trained under the CTDE paradigm because it offers a stable and scalable framework for training many agents.}
\end{figure}

\begin{figure*}[t!]
  \centering
  \includegraphics[width=\linewidth]{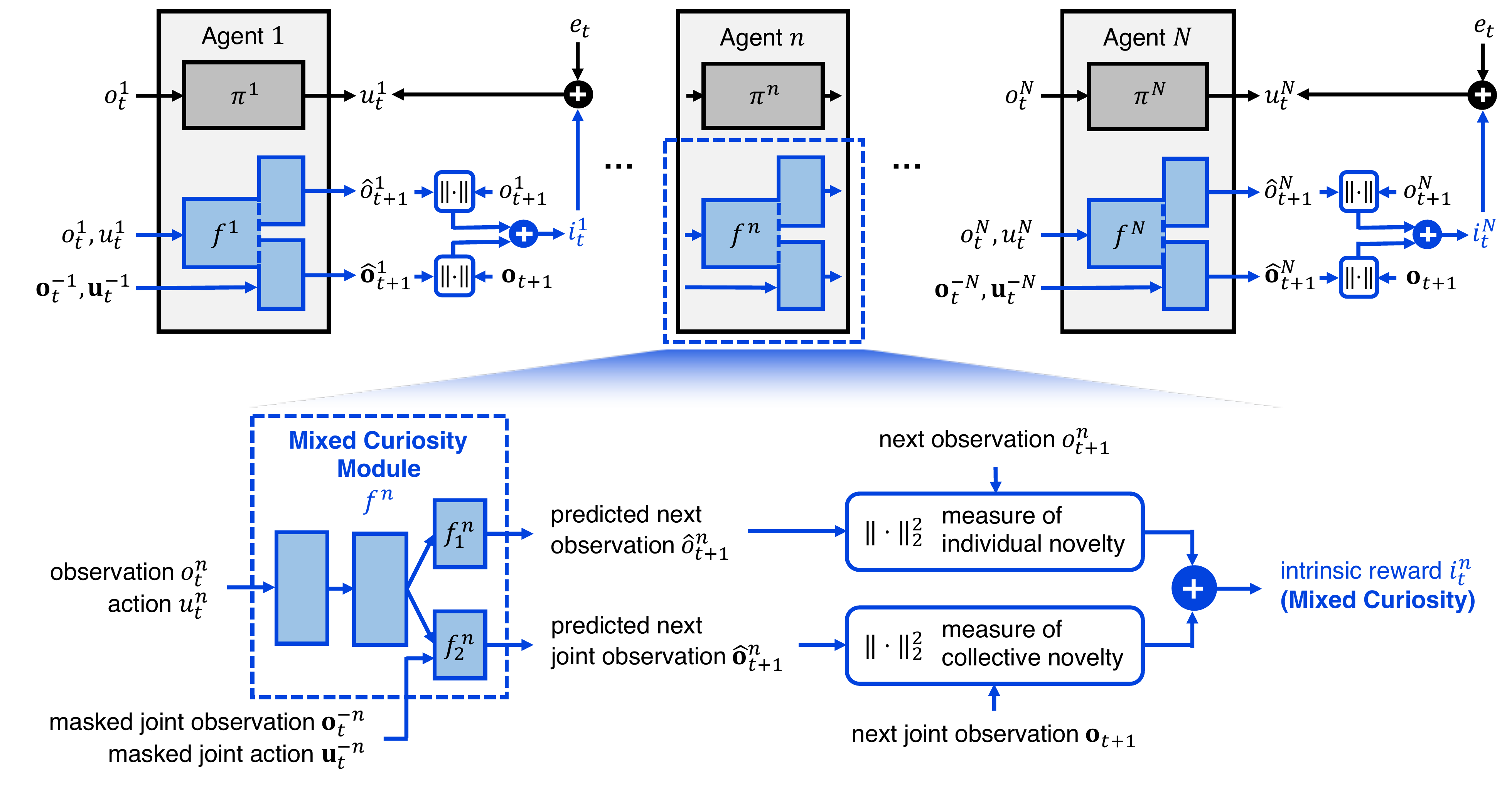}
  \caption{Our proposed curiosity-driven multi-agent exploration method. We use as an intrinsic reward the sum of the individual and joint prediction errors of our Mixed Curiosity Module in order to motivate the agents to explore the environment in ways that are individually and collectively novel.}
  \label{fig:fig3}
  \Description{Our proposed curiosity-driven multi-agent exploration method that has the mixed objective of motivating the agents to explore the environment in ways that are individually and collectively novel.}
\end{figure*}

\subsection{Cooperative Multi-Agent Reinforcement Learning}
We consider a decentralized Markov decision process (DEC-MDP) which can be described by a tuple $G=\langle A, S, O, U, P, e, \gamma \rangle$, where
\begin{itemize}
    \item $A=\{1,2,\ldots,N\}$ is the set of $N$ agents,
    \item $S$ is the global state of the environment,
    \item $O = \{O_1, O_2, \ldots, O_N\}$ is the observation space of the agents,
    \item $U = \{U_1, U_2, \ldots, U_N\}$ is the action space of the agents,
    \item $P(s_{t+1}|s_t,\textbf{u}_t): S \times U \times S \rightarrow [0,1]$ is the state transition function,
    \item $e(s_t,\textbf{u}_t):S \times U \rightarrow \mathbb{R}$ is the extrinsic reward function or the reward function of the environment, and
    \item $\gamma \in [0,1)$ is the discount factor.
\end{itemize}
At each time step $t$, each agent $n \in A$ receives its own partial observation $o_t^n \in O_n$ which is part of the joint observation $\textbf{o}_t = \{o_t^n\}_{n=1}^N$. Since this is a DEC-MDP, the global state $s_t \in S$ is uniquely defined by the joint observation $\textbf{o}_t$. Each agent $n$ also chooses its own action $u_t^n \in U_n$ which is part of the joint action $\textbf{u}_t = \{u_t^n\}_{n=1}^N$. We use the notations $\textbf{o}_t^{-n}$ and $\textbf{u}_t^{-n}$ to refer to the joint observation and joint action where agent $n$'s observation and action are respectively masked out. Given the joint observation $\textbf{o}_t$ and joint action $\textbf{u}_t$, the system transitions to the next joint observation $\textbf{o}_{t+1}$ according to state transition function $P$. Considering fully cooperative systems, all agents then receive the same extrinsic reward $e_t$. Let $\pi_n(u_t^n|o_t^n): O_n \times U_n \rightarrow [0,1]$ be the stochastic policy of agent $n$ and denote $\pi = \{\pi_1, \pi_2, \ldots, \pi_N\}$. The goal of the agents is to maximize the discounted expected return  $R_t = \sum_{\ell=0}^{\infty}\gamma^\ell e_{t+\ell}$ which is approximated by a value function $V_\pi(\textbf{o}_t)=\mathbb{E}_{\textbf{u}_t,\textbf{o}_{t+1},\ldots}\big[R_t|\textbf{o}_t\big]$ and/or an action-value function $Q_\pi(\textbf{o}_t,\textbf{u}_t)=\mathbb{E}_{\textbf{o}_{t+1},\textbf{u}_{t+1},\ldots}\big[R_t|\textbf{o}_t,\textbf{u}_t\big]$.

\subsection{Paradigm of Centralized Training with Decentralized Execution}

Works on MARL fall under three learning paradigms: (1) fully decentralized, (2) centralized training with decentralized execution (CTDE), and (3) fully centralized. Among them, the CTDE paradigm has become the popular choice of many works because the fully decentralized and fully centralized paradigms suffer from stability and scalability issues, respectively.  On the contrary, the CTDE paradigm is able to address both these problems by maintaining individual policies while reasonably assuming that they can be centrally trained with the full state information, as shown in Figure 2. It also assumes that agents can only act upon their own local information and are under some communication constraints during execution time, which is usually the case for many practical real-world scenarios.
\section{Curiosity-Driven Multi-Agent Exploration with Mixed Objectives}
In this section, we propose a new curiosity-driven method that can efficiently encourage exploration in cooperative MARL by having the mixed objective of motivating each agent to explore the environment in individually and collectively novel ways. We provide an illustration of our proposed method in Figure \ref{fig:fig3}.

\subsection{Curiosity-Driven Exploration}
We build upon the curiosity-driven approach of Pathak et al. \cite{icm} which uses the prediction error of a novelty module, called Intrinsic Curiosity Module (ICM), as an intrinsic reward to guide the exploration of a single RL agent in sparse reward environments. 

ICM is a neural network $f$ that is trained to predict the agent's next state $s_{t+1}$ given its current state $s_t$ and action $u_t$:
\begin{equation}
\hat{s}_{t+1} = f(s_t,u_t),
\end{equation}
where $\hat{s}_{t+1}$ is the predicted next state and $f$ is trained to minimize loss function $L$:
\begin{equation}
L = \|\hat{s}_{t+1} - s_{t+1}\|_2^2.    
\end{equation}
The agent then generates its intrinsic reward $i_t$ from $f$ using novelty function $g$, which is the error between the predicted next state $\hat{s}_{t+1}$ and the true next state $s_{t+1}$:
\begin{align}
\begin{split}
i_t &= g(\hat{s}_{t+1},s_{t+1})\\
&= \|\hat{s}_{t+1} - s_{t+1}\|_2^2.
\end{split}
\end{align}

Viewed as a measure of curiosity, the prediction error of a neural network quantifies the novelty of an agent's experience based on the agent's ability to predict the consequences of its actions. In the lack or even absence of rewards from the environment, curiosity has been shown to be an effective intrinsic reward signal on how the agent should explore the environment, that is to look for novel experiences in order to reduce its uncertainty of how the world works \cite{icm,large-scale}. Recent works inspired by ICM designed new curiosity module architectures and intrinsic reward formulations to improve the performance of ICM in stochastic \cite{disagreement,rnd,lwm} and procedurally-generated \cite{ride} environments, but these methods were still designed for single-agent scenarios. Here, we propose an effective curiosity-driven exploration method for cooperative MARL.

\subsection{Mixed Curiosity as an Intrinsic Reward}
Similar to single-agent RL algorithms, the performance of cooperative MARL algorithms also degrades in sparse reward environments (see Figure \ref{fig:fig1}). As is done in the single-agent case, we can straightforwardly equip each agent with its own ICM and use the prediction error of this curiosity module as the agent's intrinsic reward. Now that we are considering a multi-agent system, we can view this individual prediction error as a measure of individual novelty since it only quantifies the gap in one agent's knowledge of the environment and disregards what other agents may already know about it. For multi-agent tasks which require cooperation among many agents, individual novelty alone as an intrinsic reward may not always be a helpful exploration bonus because it will motivate the agents to explore the environment independently. Such an exploration strategy is simply inefficient.

As explained in \cite{coordinated_exploration}, a better intrinsic reward signal in the multi-agent case is a measure of collective novelty. For example, we can take the minimum among the set of individual prediction errors \cite{coordinated_exploration} of per-agent curiosity modules or use the joint prediction error of a joint curiosity module, a neural network that predicts the next joint observation given the current joint observation and joint action \cite{synergy}. Naturally, a measure of collective novelty will induce cooperation because it is based on the group's combined knowledge. However, it may likewise be just another vague signal to the agents because they may find it difficult to identify from it how well each of them behaved, another challenge in MARL known as the credit assignment problem. 

Despite these, we show here that curiosity can still be an effective intrinsic reward signal in the multi-agent case by making simple but effective modifications to the ICM framework. Our main contribution is a new curiosity-driven multi-agent exploration method that has the mixed objective of motivating the agents to explore the environment in individually and collectively novel ways. In line with this, we develop a two-headed neural network called Mixed Curiosity Module (MCM) and design the intrinsic reward formula to be the sum of the individual and joint prediction errors of MCM. We demonstrate later in our experiments that compared to other baselines, our proposed method provides the best performance boost to cooperative MARL algorithms in sparse reward environments.
\\

\noindent \textbf{Curiosity Module Architecture.} Each agent $n$ is equipped with an MCM $f^n$ that is trained to predict agent $n$'s individual next observation $o_{t+1}^n$ in the first head $f^n_1$ and the next joint observation $\textbf{o}_{t+1}$ in the second head $f^n_2$:
\begin{align}
\hat{o}_{t+1}^{n} &= f^n_1(o_t^n,u_t^n),\\
\hat{\textbf{o}}_{t+1}^{n} &= f^n_2(o_t^n,u_t^n,\textbf{o}_t^{-n},\textbf{u}_t^{-n}),
\end{align}
where $\hat{o}_{t+1}^{n}$ is the predicted next observation of agent $n$, $\hat{\textbf{o}}_{t+1}^{n}$ is the predicted next joint observation, and $f^n$ is trained to minimize loss function $L^n$: 
\begin{equation}
L^n = \frac{1}{2}(\|\hat{o}_{t+1}^{n} - o_{t+1}^n\|_2^2 + \|\hat{\textbf{o}}_{t+1}^{n} - \textbf{o}_{t+1}\|_2^2).    
\end{equation}
Since predicting the next joint observation will require information from all agents, we give the observation and action of all other agents as input to the second head. Note that this is only passed to the second head because predicting the next observation of the corresponding agent does not need this additional information.
\\

\noindent \textbf{Intrinsic Reward Formulation.} Every agent $n$ generates its intrinsic reward $i_t^n$ from $f^n$ using novelty function $g^n$:
\begin{align}
\begin{split}
i_t^{n} &= g^n(\hat{o}^n_{t+1},\hat{\textbf{o}}^n_{t+1},\textbf{o}_{t+1})\\
&= \|\hat{o}_{t+1}^{n} - o_{t+1}^n\|_2^2 + \|\hat{\textbf{o}}_{t+1}^{n} - \textbf{o}_{t+1}\|_2^2,    
\end{split}
\end{align}
where the first and second terms are the individual and joint prediction errors of agent $n$'s MCM, respectively. We can accordingly view these prediction errors as measures of individual and collective novelties. Thus, we consider our intrinsic reward formulation to be a form of mixed curiosity.

\section{Experiments}
We evaluate our approach on two sparse reward cooperative navigation scenarios and compare the results against existing works and ablations of our method.

\subsection{Experimental Setup}
\noindent \textbf{Environment.} We run our experiments on the cooperative navigation environment in \cite{particle_environment}. In this environment, $N$ agents and $L$ landmarks inhabit a bounded two-dimensional world with continuous state space. The $N$ agents are tasked to reach a set of $L$ landmarks while avoiding collisions with other agents. Each agent receives as observation the relative positions of the landmarks and the other agents. They are also able to select from a discrete set of actions: \{up, down, left, right, stay\}. Usually, this environment is used with a dense reward function where agents are collectively rewarded based on the proximity of any agent to each landmark. Additionally, they also receive a penalty for colliding with other agents and for going out of the allotted dimensions of the world. To make the rewards sparse, we modify the reward function such that the agents only obtain an extrinsic reward when all of them are simultaneously within small distances from their assigned landmarks. We observed that the performance of cooperative MARL algorithms already degrades in this setup. 
\\

\begin{figure}[h!]
  \centering
  \includegraphics[scale=0.45]{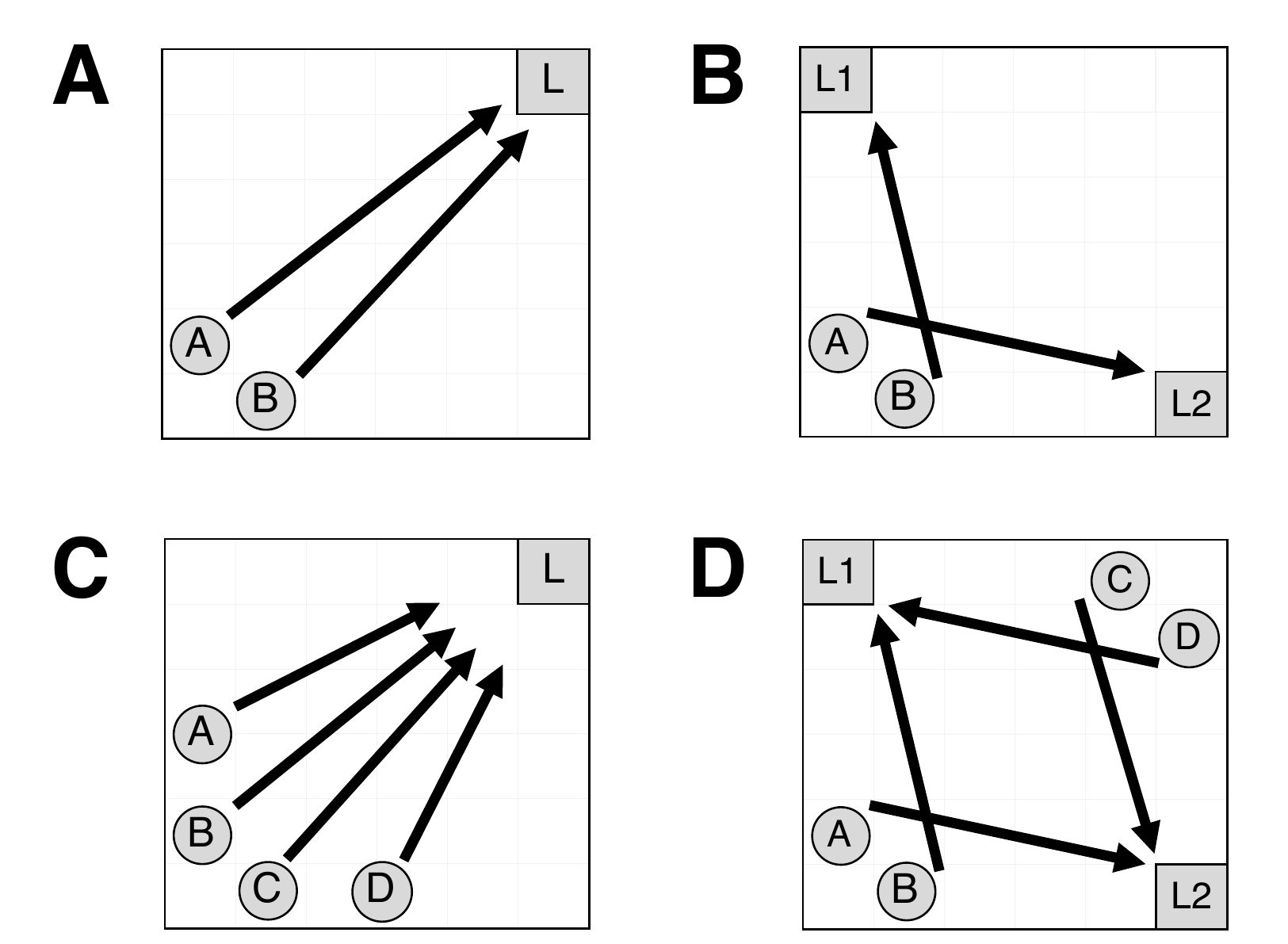}
  \caption{Cooperative navigation scenarios: (A) 2-agent same landmark scenario, (B) 2-agent different landmark scenario, (C) 4-agent same landmark scenario, and (D) 4-agent different landmark scenario.}
  \label{fig:fig4}
  \Description{Cooperative navigation scenarios: (A) 2-agent same landmark scenario, (B) 2-agent different landmark scenario, (C) 4-agent same landmark scenario, and (D) 4-agent different landmark scenario.}
\end{figure}

\noindent \textbf{Scenarios.} We consider two scenarios of the cooperative navigation environment: (1) with same landmark and (2) with different landmarks. In the same landmark scenario, all agents must simultaneously reach one landmark which is located at the farthest position on the map from the agents' starting positions. Whereas in the different landmark scenario, agents are tasked to reach two different landmarks which are located at opposite positions on the map. We provide in Figure \ref{fig:fig4} a visualization of the 2-agent and 4-agent versions of these scenarios. 

\begin{table*}[t!]
\centering
\caption{A side-by-side comparison of the curiosity module architecture and intrinsic reward formulation of the curiosity-driven multi-agent exploration methods that we considered in our experiments.}
\label{tab:baselines}
\begin{tabular}{@{}llll@{}}
\toprule
\textbf{Method}    
& \textbf{Curiosity Module Architecture} 
& {\textbf{Intrinsic Reward Formulation}}
& {\textbf{Reference}}                              
\\ 
\midrule
COMA&None&None&\cite{coma}\\
COMA+ICM-Indiv&$N$ one-headed curiosity modules&Individual prediction error&\cite{coordinated_exploration,synergy}\\
COMA+ICM-Joint&$1$ one-headed curiosity module&Joint prediction error&\cite{synergy}\\
COMA+ICM-Min&$N$ one-headed curiosity modules&Minimum of individual prediction errors&\cite{coordinated_exploration}\\
\midrule
COMA+MCM-Indiv&$N$ two-headed curiosity modules&Individual prediction error&Ablation\\
COMA+MCM-Joint&$N$ two-headed curiosity modules&Joint prediction error&Ablation\\
COMA+MCM-Sep&$N$$+$$1$ one-headed curiosity modules&Sum of individual prediction error and joint prediction error&Ablation\\
\midrule
COMA+MCM&$N$ two-headed curiosity modules&Sum of individual prediction error and joint prediction error&Ours\\
\bottomrule
\end{tabular}
\end{table*}

\subsection{Implementation Details}
\noindent \textbf{Network Architecture.} All baselines make use of a standard curiosity module $f^n$ and a policy network $\pi^n$ which are 3-layer fully connected neural networks with 64-unit hidden layers and leaky ReLU activation functions. We note that in our earlier experiments, we used the full architecture of ICM which has a forward model, an inverse model, and state encoders \cite{icm}. The forward model is the one used to generate the intrinsic rewards, whereas the inverse model and the state encoders are primarily used for feature encoding. However, we observed that learning with this architecture is too slow in our cooperative navigation environment. We hypothesized this is because we are doing unnecessary preprocessing of inputs. The environments used in \cite{icm} have high dimensional images as inputs. These images carry many irrelevant information which can be reduced with the help of the inverse model and state encoders. However, this is not the case in most multi-agent environments, such as the cooperative navigation environment, where the inputs already have a compact representation (see Section 5.1). we verified that this is the case by using an architecture with the forward model only and observed better results. So in this paper, we ignored the inverse model and the state encoders in all discussions about ICM and in our experiments as well.
\\ 

\noindent \textbf{Training Details.} We optimize all forward models and policy networks concurrently from scratch. As our baseline algorithm, we use COMA \cite{coma}. We implement all methods using the publicly available code in \cite{maac} and run them with the default settings. We clip the intrinsic rewards to 1 and set the trade-off coefficient $\lambda$ to 0.05.

\subsection{Baselines}
\noindent \textbf{Performance Baselines.} We compare the performance of our proposed method against other curiosity-driven exploration methods which we adapted from related works.
\begin{itemize}
    \item COMA: This baseline is the original algorithm. It does not use any intrinsic rewards for exploration.
    \item COMA+ICM-Indiv: Each agent $n$ uses as an intrinsic reward the individual prediction error of its ICM $f^n$.
    \item COMA+ICM-Joint: Each agent $n$ uses as an intrinsic reward the joint prediction error of a joint ICM that is shared by all agents. Unlike other methods, this baseline provides all agents with the same intrinsic reward values.
    \item COMA+ICM-Min: Each agent $n$ uses as an intrinsic reward the minimum among the individual prediction errors of each agent's ICM on agent $n$'s current state and action. For the exact definition, we kindly refer the reader to \cite{coordinated_exploration}.
    \item COMA+MCM: Our proposed method.\vspace{5pt}
\end{itemize}
\noindent \textbf{Ablation Study Baselines.} We also provide evaluations against ablations of our proposed method:
\begin{itemize}
    \item COMA+MCM-Indiv: Each agent $n$ uses as an intrinsic reward only the individual prediction error of its MCM $f^n$.
    \item COMA+MCM-Joint: Each agent $n$ uses as an intrinsic reward only the joint prediction error of its MCM $f^n$.
    \item COMA+MCM-Sep: Each agent $n$ uses as an intrinsic reward the sum of the individual prediction error of its ICM $f^n$ and the joint prediction error of a separate joint ICM. 
\end{itemize}

\noindent We provide a summary of these baselines in Table \ref{tab:baselines}.  
\begin{table*}[t!]
\centering
\caption{Final results of the performance baselines on the cooperative navigation scenarios. Best results are written in bold.}
\label{tab:performance}
\begin{tabular}{@{}llllll@{}}
\toprule
\textbf{Method}    
& \multicolumn{2}{l}{\textbf{Same Landmark}} 
& \multicolumn{2}{l}{\textbf{Different Landmark}}
& \textbf{Reference}
\\
\cmidrule(l){2-5} & 2-agent & 4-agent & 2-agent & 4-agent
\\
\midrule
COMA&$0.53\pm0.23$&$0.41\pm0.28$&$0.45\pm0.23$&$0.31\pm0.28$&\cite{coma}\\
COMA+ICM-Indiv&$0.80\pm0.16$&$\boldsymbol{0.68\pm0.25}$&$0.47\pm0.24$&$0.16\pm0.23$&\cite{synergy,coordinated_exploration}\\
COMA+ICM-Joint&$0.69\pm0.20$&$0.31\pm0.27$&$0.52\pm0.22$&$0.12\pm0.18$&\cite{synergy}\\
COMA+ICM-Min&$0.57\pm0.24$&$0.00\pm0.00$&$0.76\pm0.19$&$0.53\pm0.32$&\cite{coordinated_exploration}\\
\midrule
COMA+MCM& $\boldsymbol{0.89\pm0.04}$&$\boldsymbol{0.68\pm0.25}$&$\boldsymbol{0.82\pm0.14}$&$\boldsymbol{0.66\pm0.25}$&Ours\\
\bottomrule
\end{tabular}
\end{table*}

\begin{figure*}[t!]
  \centering
  \includegraphics[width=\linewidth]{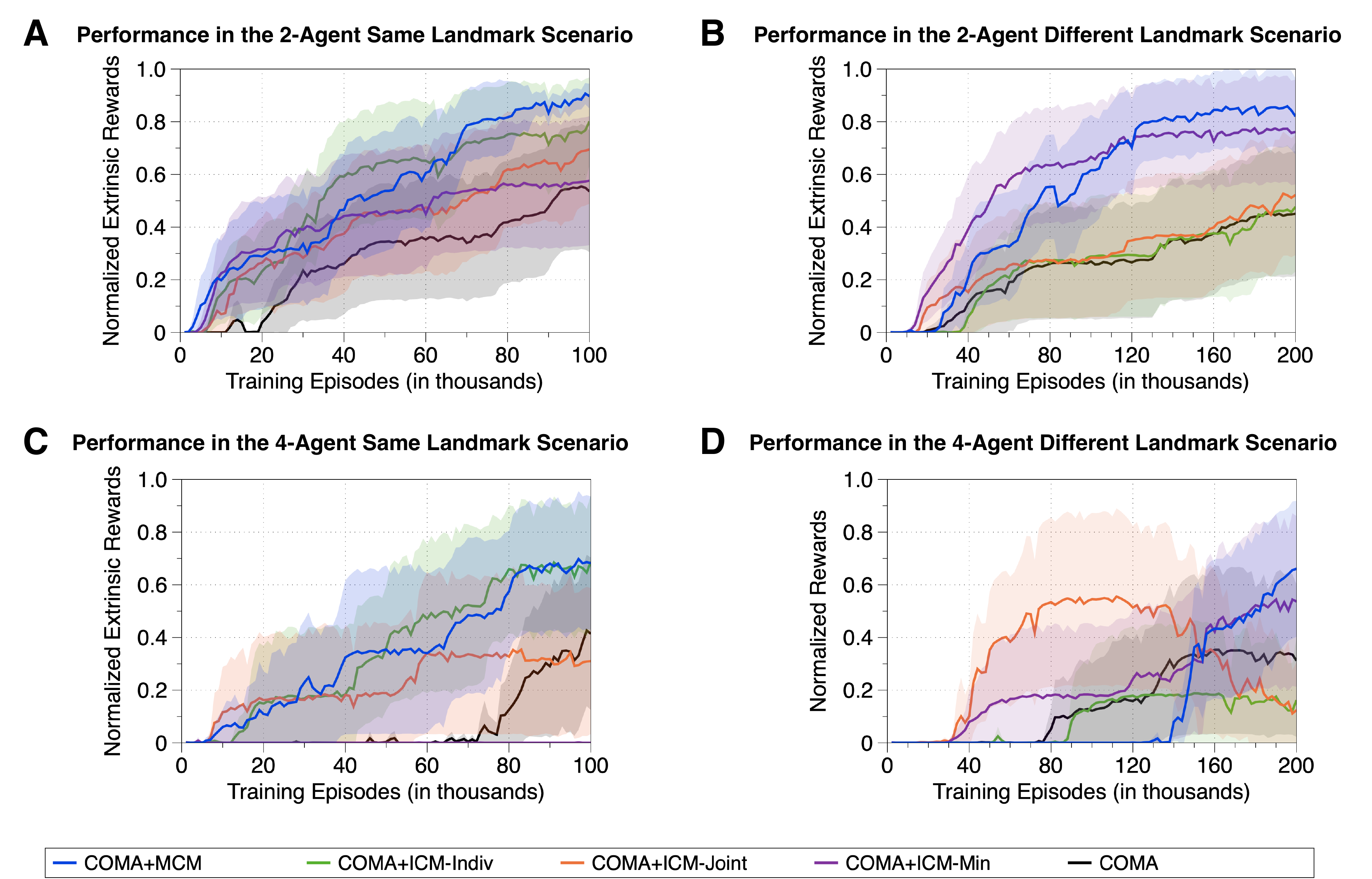}
  \caption{Performance results: Normalized mean episode rewards on the cooperative navigation environments with 90\% confidence intervals, using 10 seeds for the 2-agent scenarios and 5 seeds for the 4-agent scenarios. These results show that the best performing method is our method, which are the blue lines in the plots.}
  \label{fig:fig5}
  \Description{Performance results: Normalized mean episode rewards on the cooperative navigation environments with 90\% confidence intervals, across 10 seeds for the 2-agent scenarios and across 5 seeds for the 4-agent scenarios. These results show that the best performing method is our method which are the blue lines in the plots.}
\end{figure*}

\begin{figure*}[t!]
  \centering
  \includegraphics[width=\linewidth]{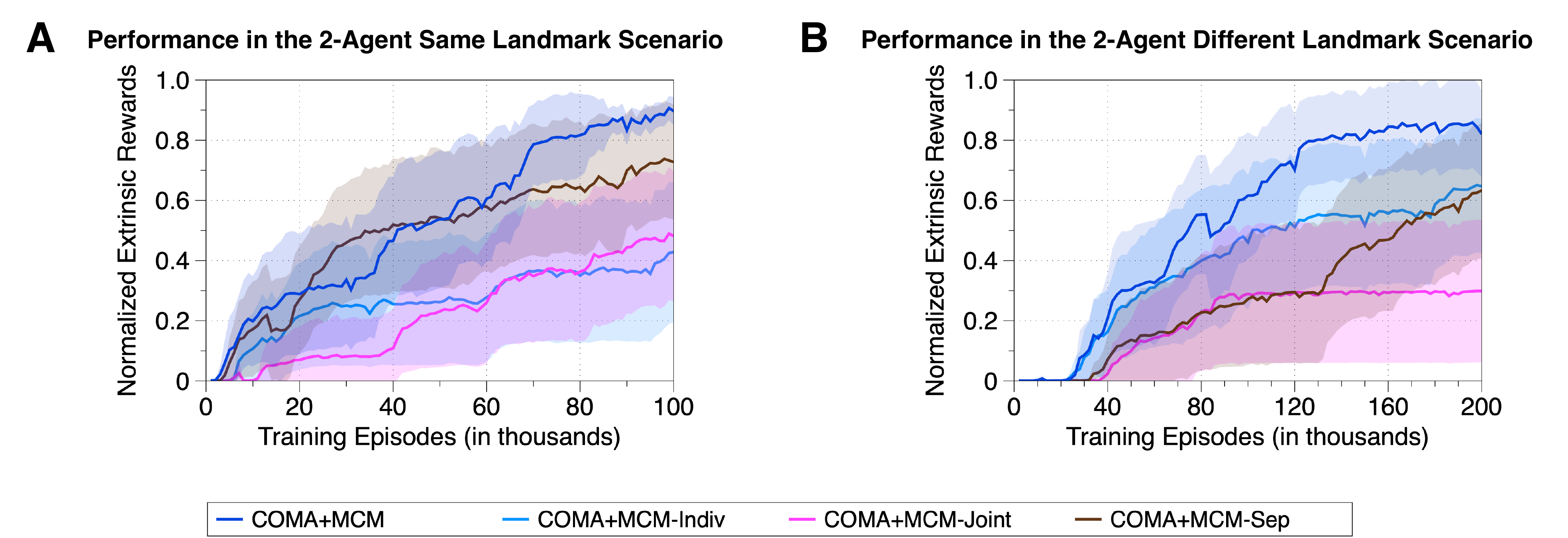}
  \caption{Ablation study results: Normalized mean episode rewards on the cooperative navigation environments with 90\% confidence intervals.}
  \label{fig:fig6}
  \Description{Ablation study results.}
\end{figure*}

\subsection{Results and Discussion}
\noindent \textbf{Performance Results.} As shown in Figure \ref{fig:fig5}, MCM gives the best performance boost to COMA in both the same landmark and different landmark scenarios compared to all performance baselines. This demonstrates that MCM provides the most effective way of encouraging exploration in cooperative MARL. 

Notably, only our approach performs well in both cooperative navigation scenarios. We can see that although COMA+ICM-Indiv and COMA+ICM-Joint perform well in the same landmark scenario, they perform poorly in the the different landmark scenario. Conversely, COMA+ICM-Min has a good performance in the different landmark scenario but not in the same landmark scenario. These results highlight the benefit of having an intrinsic reward signal which incorporates both measures of individual and collective novelties. There are situations where individual novelty alone serves as a sufficient intrinsic reward signal. This happens when agents can independently explore the environment but still solve the task at hand, such as the case of the same landmark scenario. We can say that using collective novelty alone in these types of situations only makes learning more difficult for the agents, as demonstrated by the poor performance of COMA+ICM-Joint in the same landmark scenario. However, there are also cases like the different landmark scenario where collective novelty serves as a better intrinsic reward signal. In such cases, individual novelty alone does not help improve performance. We can see that COMA+ICM-Indiv has about the same or even worse performance than COMA in the different landmark scenario. Unlike these na\"ive baselines, our proposed method is robust to different types of multi-agent situations. We also note that these observations are consistent in both the 2-agent and 4-agent versions of the two cooperative navigation scenarios.
\\

\begin{table}[t!]
\centering
\caption{Final results of the ablation baselines on the 2-agent version of both cooperative navigation scenarios. Best results are written in bold.}
\label{tab:ablation}
\begin{tabular}{@{}lll@{}}
\toprule
\textbf{Method}    
& \multicolumn{2}{l}{\textbf{Cooperative Navigation Scenario}}
\\
\cmidrule(l){2-3} & Same Landmark & Different Landmark
\\
\midrule
COMA+MCM-Indiv&$0.42\pm0.23$&$0.64\pm0.22$\\
COMA+MCM-Joint&$0.48\pm0.21$&$0.29\pm0.23$\\
COMA+MCM-Sep&$0.72\pm0.19$&$0.63\pm0.22$\\
\midrule
COMA+MCM& $\boldsymbol{0.89\pm0.04}$&$\boldsymbol{0.68\pm0.25}$\\
\bottomrule
\end{tabular}
\end{table}

\noindent \textbf{Ablation Study Results.} We additionally conducted ablation studies to see how important is each of our contributions. To test the effectiveness of our curiosity module architecture, we compare COMA+MCM against COMA+MCM-Sep which still uses the sum of the individual and joint prediction errors as an intrinsic reward but those errors are generated by separate ICMs, one predicts just the corresponding agent's next observation and the other predicts the joint next observation only. Moreover, we consider COMA+MCM-Indiv and COMA+MCM-Joint to evaluate our proposed intrinsic reward formulation. These two ablations equip the agents with MCM but they only use its individual prediction error and joint prediction error, respectively. As we can see in Figure \ref{fig:fig6}, all these ablation baselines have worse performances than our method. Thus, these results show that our proposed curiosity module architecture and intrinsic reward formulation are both important contributions.
\section{Conclusion}
Curiosity is among the intrinsic reward signals that can efficiently encourage exploration in single-agent environments with sparse rewards. In this paper, we showed that with simple but novel modifications, curiosity can also effectively motivate exploration in cooperative MARL. In particular, we developed a Mixed Curiosity Module (MCM) which is a two-headed neural network that is trained to predict the corresponding agent's next observation in the first head and the next joint observation in the second head. Moreover, we designed an intrinsic reward formulation that incorporates measures of both individual and collective novelties by using the sum of the individual and joint prediction errors of each agent's MCM as the agent's intrinsic reward. We demonstrated in two cooperative navigation scenarios that our proposed method, which has the mixed objective of motivating the agents to explore the environment in individually and collectively novel ways, provides the best performance boost to standard cooperative MARL algorithms. For future work, we plan to test our method on more sparse reward multi-agent environments.

\bibliographystyle{ACM-Reference-Format} 
\bibliography{main}


\begin{thebibliography}{30}


\ifx \showCODEN    \undefined \def \showCODEN     #1{\unskip}     \fi
\ifx \showDOI      \undefined \def \showDOI       #1{#1}\fi
\ifx \showISBNx    \undefined \def \showISBNx     #1{\unskip}     \fi
\ifx \showISBNxiii \undefined \def \showISBNxiii  #1{\unskip}     \fi
\ifx \showISSN     \undefined \def \showISSN      #1{\unskip}     \fi
\ifx \showLCCN     \undefined \def \showLCCN      #1{\unskip}     \fi
\ifx \shownote     \undefined \def \shownote      #1{#1}          \fi
\ifx \showarticletitle \undefined \def \showarticletitle #1{#1}   \fi
\ifx \showURL      \undefined \def \showURL       {\relax}        \fi
\providecommand\bibfield[2]{#2}
\providecommand\bibinfo[2]{#2}
\providecommand\natexlab[1]{#1}
\providecommand\showeprint[2][]{arXiv:#2}

\bibitem[\protect\citeauthoryear{Anonymous}{Anonymous}{2021}]%
        {cmae}
\bibfield{author}{\bibinfo{person}{Anonymous}.}
  \bibinfo{year}{2021}\natexlab{}.
\newblock \showarticletitle{Coordinated Multi-Agent Exploration Using Shared
  Goals}. In \bibinfo{booktitle}{\emph{Submitted to ICLR}}.
\newblock
\newblock
\shownote{Under review.}


\bibitem[\protect\citeauthoryear{Bellemare, Srinivasan, Ostrovski, Schaul,
  Saxton, and Munos}{Bellemare et~al\mbox{.}}{2016}]%
        {unifying}
\bibfield{author}{\bibinfo{person}{Marc~G. Bellemare}, \bibinfo{person}{Sriram
  Srinivasan}, \bibinfo{person}{Georg Ostrovski}, \bibinfo{person}{Tom Schaul},
  \bibinfo{person}{David Saxton}, {and} \bibinfo{person}{R{\'{e}}mi Munos}.}
  \bibinfo{year}{2016}\natexlab{}.
\newblock \showarticletitle{Unifying Count-Based Exploration and Intrinsic
  Motivation}. In \bibinfo{booktitle}{\emph{Proceedings of NeurIPS}}.
\newblock


\bibitem[\protect\citeauthoryear{B{\"{o}}hmer, Rashid, and
  Whiteson}{B{\"{o}}hmer et~al\mbox{.}}{2019}]%
        {unreliable}
\bibfield{author}{\bibinfo{person}{Wendelin B{\"{o}}hmer},
  \bibinfo{person}{Tabish Rashid}, {and} \bibinfo{person}{Shimon Whiteson}.}
  \bibinfo{year}{2019}\natexlab{}.
\newblock \showarticletitle{Exploration with Unreliable Intrinsic Reward in
  Multi-Agent Reinforcement Learning}.
\newblock \bibinfo{journal}{\emph{arXiv preprint arXiv:1906.02138}}
  (\bibinfo{year}{2019}).
\newblock


\bibitem[\protect\citeauthoryear{Burda, Edwards, Pathak, Storkey, Darrell, and
  Efros}{Burda et~al\mbox{.}}{2019a}]%
        {large-scale}
\bibfield{author}{\bibinfo{person}{Yuri Burda}, \bibinfo{person}{Harrison
  Edwards}, \bibinfo{person}{Deepak Pathak}, \bibinfo{person}{Amos~J. Storkey},
  \bibinfo{person}{Trevor Darrell}, {and} \bibinfo{person}{Alexei~A. Efros}.}
  \bibinfo{year}{2019}\natexlab{a}.
\newblock \showarticletitle{Large-Scale Study of Curiosity-Driven Learning}. In
  \bibinfo{booktitle}{\emph{Proceedings of ICLR}}.
\newblock


\bibitem[\protect\citeauthoryear{Burda, Edwards, Storkey, and Klimov}{Burda
  et~al\mbox{.}}{2019b}]%
        {rnd}
\bibfield{author}{\bibinfo{person}{Yuri Burda}, \bibinfo{person}{Harrison
  Edwards}, \bibinfo{person}{Amos~J. Storkey}, {and} \bibinfo{person}{Oleg
  Klimov}.} \bibinfo{year}{2019}\natexlab{b}.
\newblock \showarticletitle{Exploration by Random Network Distillation}. In
  \bibinfo{booktitle}{\emph{Proceedings of ICLR}}.
\newblock


\bibitem[\protect\citeauthoryear{Chitnis, Tulsiani, Gupta, and Gupta}{Chitnis
  et~al\mbox{.}}{2020}]%
        {synergy}
\bibfield{author}{\bibinfo{person}{Rohan Chitnis}, \bibinfo{person}{Shubham
  Tulsiani}, \bibinfo{person}{Saurabh Gupta}, {and} \bibinfo{person}{Abhinav
  Gupta}.} \bibinfo{year}{2020}\natexlab{}.
\newblock \showarticletitle{Intrinsic Motivation for Encouraging Synergistic
  Behavior}. In \bibinfo{booktitle}{\emph{Proceedings of ICLR}}.
\newblock


\bibitem[\protect\citeauthoryear{Ermolov and Sebe}{Ermolov and Sebe}{2020}]%
        {lwm}
\bibfield{author}{\bibinfo{person}{Aleksandr Ermolov} {and}
  \bibinfo{person}{Nicu Sebe}.} \bibinfo{year}{2020}\natexlab{}.
\newblock \showarticletitle{Latent World Models For Intrinsically Motivated
  Exploration}. In \bibinfo{booktitle}{\emph{Proceedings of NeurIPS}}.
\newblock


\bibitem[\protect\citeauthoryear{Eysenbach, Gupta, Ibarz, and Levine}{Eysenbach
  et~al\mbox{.}}{2019}]%
        {diversity}
\bibfield{author}{\bibinfo{person}{Benjamin Eysenbach},
  \bibinfo{person}{Abhishek Gupta}, \bibinfo{person}{Julian Ibarz}, {and}
  \bibinfo{person}{Sergey Levine}.} \bibinfo{year}{2019}\natexlab{}.
\newblock \showarticletitle{Diversity is All You Need: Learning Skills without
  a Reward Function}. In \bibinfo{booktitle}{\emph{Proceedings of ICLR}}.
\newblock


\bibitem[\protect\citeauthoryear{Foerster, Farquhar, Afouras, Nardelli, and
  Whiteson}{Foerster et~al\mbox{.}}{2018}]%
        {coma}
\bibfield{author}{\bibinfo{person}{Jakob~N. Foerster}, \bibinfo{person}{Gregory
  Farquhar}, \bibinfo{person}{Triantafyllos Afouras}, \bibinfo{person}{Nantas
  Nardelli}, {and} \bibinfo{person}{Shimon Whiteson}.}
  \bibinfo{year}{2018}\natexlab{}.
\newblock \showarticletitle{Counterfactual Multi-Agent Policy Gradients}. In
  \bibinfo{booktitle}{\emph{Proceedings of {AAAI}}}.
\newblock


\bibitem[\protect\citeauthoryear{Houthooft, Chen, Chen, Duan, Schulman,
  De~Turck, and Abbeel}{Houthooft et~al\mbox{.}}{2016}]%
        {vime}
\bibfield{author}{\bibinfo{person}{Rein Houthooft}, \bibinfo{person}{Xi Chen},
  \bibinfo{person}{Xi Chen}, \bibinfo{person}{Yan Duan}, \bibinfo{person}{John
  Schulman}, \bibinfo{person}{Filip De~Turck}, {and} \bibinfo{person}{Pieter
  Abbeel}.} \bibinfo{year}{2016}\natexlab{}.
\newblock \showarticletitle{VIME: Variational Information Maximizing
  Exploration}. In \bibinfo{booktitle}{\emph{Proceedings of NeurIPS}}.
\newblock


\bibitem[\protect\citeauthoryear{Iqbal and Sha}{Iqbal and Sha}{2019a}]%
        {maac}
\bibfield{author}{\bibinfo{person}{Shariq Iqbal} {and} \bibinfo{person}{Fei
  Sha}.} \bibinfo{year}{2019}\natexlab{a}.
\newblock \showarticletitle{Actor-Attention-Critic for Multi-Agent
  Reinforcement Learning}. In \bibinfo{booktitle}{\emph{Proceedings of ICML}}.
\newblock


\bibitem[\protect\citeauthoryear{Iqbal and Sha}{Iqbal and Sha}{2019b}]%
        {coordinated_exploration}
\bibfield{author}{\bibinfo{person}{Shariq Iqbal} {and} \bibinfo{person}{Fei
  Sha}.} \bibinfo{year}{2019}\natexlab{b}.
\newblock \showarticletitle{Coordinated Exploration via Intrinsic Rewards for
  Multi-Agent Reinforcement Learning}.
\newblock \bibinfo{journal}{\emph{arXiv preprint arXiv:1905.12127}}
  (\bibinfo{year}{2019}).
\newblock


\bibitem[\protect\citeauthoryear{Jiang and Lu}{Jiang and Lu}{2020}]%
        {eoi}
\bibfield{author}{\bibinfo{person}{Jiechuan Jiang} {and}
  \bibinfo{person}{Zongqing Lu}.} \bibinfo{year}{2020}\natexlab{}.
\newblock \showarticletitle{The Emergence of Individuality in Multi-Agent
  Reinforcement Learning}.
\newblock \bibinfo{journal}{\emph{arXiv preprint arXiv:2006.05842}}
  (\bibinfo{year}{2020}).
\newblock


\bibitem[\protect\citeauthoryear{Kim, Kim, Jeong, Levine, and Song}{Kim
  et~al\mbox{.}}{2019}]%
        {emi}
\bibfield{author}{\bibinfo{person}{Hyoungseok Kim}, \bibinfo{person}{Jaekyeom
  Kim}, \bibinfo{person}{Yeonwoo Jeong}, \bibinfo{person}{Sergey Levine}, {and}
  \bibinfo{person}{Hyun~Oh Song}.} \bibinfo{year}{2019}\natexlab{}.
\newblock \showarticletitle{{EMI:} Exploration with Mutual Information
  Maximizing State and Action Embeddings}. In
  \bibinfo{booktitle}{\emph{Proceedings of ICML}}.
\newblock


\bibitem[\protect\citeauthoryear{Lowe, Wu, Tamar, Harb, Pieter, and
  Mordatch}{Lowe et~al\mbox{.}}{2017}]%
        {maddpg}
\bibfield{author}{\bibinfo{person}{Ryan Lowe}, \bibinfo{person}{Yi Wu},
  \bibinfo{person}{Aviv Tamar}, \bibinfo{person}{Jean Harb},
  \bibinfo{person}{Abbeel Pieter}, {and} \bibinfo{person}{Igor Mordatch}.}
  \bibinfo{year}{2017}\natexlab{}.
\newblock \showarticletitle{Multi-Agent Actor-Critic for Mixed
  Cooperative-Competitive Environments}. In
  \bibinfo{booktitle}{\emph{Proceedings of NeurIPS}}.
\newblock


\bibitem[\protect\citeauthoryear{Mahajan, Rashid, Samvelyan, and
  Whiteson}{Mahajan et~al\mbox{.}}{2019}]%
        {maven}
\bibfield{author}{\bibinfo{person}{Anuj Mahajan}, \bibinfo{person}{Tabish
  Rashid}, \bibinfo{person}{Mikayel Samvelyan}, {and} \bibinfo{person}{Shimon
  Whiteson}.} \bibinfo{year}{2019}\natexlab{}.
\newblock \showarticletitle{MAVEN: Multi-Agent Variational Exploration}. In
  \bibinfo{booktitle}{\emph{Proceedings of NeurIPS}}.
\newblock


\bibitem[\protect\citeauthoryear{Mordatch and Abbeel}{Mordatch and
  Abbeel}{2017}]%
        {particle_environment}
\bibfield{author}{\bibinfo{person}{Igor Mordatch} {and} \bibinfo{person}{Pieter
  Abbeel}.} \bibinfo{year}{2017}\natexlab{}.
\newblock \showarticletitle{Emergence of Grounded Compositional Language in
  Multi-Agent Populations}.
\newblock \bibinfo{journal}{\emph{arXiv preprint arXiv:1703.04908}}
  (\bibinfo{year}{2017}).
\newblock


\bibitem[\protect\citeauthoryear{Ostrovski, Bellemare, van~den Oord, and
  Munos}{Ostrovski et~al\mbox{.}}{2017}]%
        {neural_density_models}
\bibfield{author}{\bibinfo{person}{Georg Ostrovski}, \bibinfo{person}{Marc~G.
  Bellemare}, \bibinfo{person}{A{\"{a}}ron van~den Oord}, {and}
  \bibinfo{person}{R{\'{e}}mi Munos}.} \bibinfo{year}{2017}\natexlab{}.
\newblock \showarticletitle{Count-Based Exploration with Neural Density
  Models}. In \bibinfo{booktitle}{\emph{Proceedings of ICML}}.
\newblock


\bibitem[\protect\citeauthoryear{Oudeyer and Kaplan}{Oudeyer and
  Kaplan}{2009}]%
        {oudeyer_intrinsic}
\bibfield{author}{\bibinfo{person}{Pierre-Yves Oudeyer} {and}
  \bibinfo{person}{Frederic Kaplan}.} \bibinfo{year}{2009}\natexlab{}.
\newblock \showarticletitle{What is intrinsic motivation? A typology of
  computational approaches}.
\newblock \bibinfo{journal}{\emph{Frontiers in Neurorobotics}}
  \bibinfo{volume}{1} (\bibinfo{year}{2009}), \bibinfo{pages}{6}.
\newblock
\showISSN{1662-5218}


\bibitem[\protect\citeauthoryear{Pathak, Agrawal, Efros, and Darrell}{Pathak
  et~al\mbox{.}}{2017}]%
        {icm}
\bibfield{author}{\bibinfo{person}{Deepak Pathak}, \bibinfo{person}{Pulkit
  Agrawal}, \bibinfo{person}{Alexei~A. Efros}, {and} \bibinfo{person}{Trevor
  Darrell}.} \bibinfo{year}{2017}\natexlab{}.
\newblock \showarticletitle{Curiosity-driven Exploration by Self-supervised
  Prediction}. In \bibinfo{booktitle}{\emph{Proceedings of ICML}}.
\newblock


\bibitem[\protect\citeauthoryear{Pathak, Gandhi, and Gupta}{Pathak
  et~al\mbox{.}}{2019}]%
        {disagreement}
\bibfield{author}{\bibinfo{person}{Deepak Pathak}, \bibinfo{person}{Dhiraj
  Gandhi}, {and} \bibinfo{person}{Abhinav Gupta}.}
  \bibinfo{year}{2019}\natexlab{}.
\newblock \showarticletitle{Self-Supervised Exploration via Disagreement}. In
  \bibinfo{booktitle}{\emph{Proceedings of ICML}}.
\newblock


\bibitem[\protect\citeauthoryear{Raileanu and Rockt{\"{a}}schel}{Raileanu and
  Rockt{\"{a}}schel}{2020}]%
        {ride}
\bibfield{author}{\bibinfo{person}{Roberta Raileanu} {and} \bibinfo{person}{Tim
  Rockt{\"{a}}schel}.} \bibinfo{year}{2020}\natexlab{}.
\newblock \showarticletitle{RIDE: Rewarding Impact-Driven Exploration for
  Procedurally-Generated Environments}. In
  \bibinfo{booktitle}{\emph{Proceedings of ICLR}}.
\newblock


\bibitem[\protect\citeauthoryear{Rashid, Samvelyan, de~Witt, Farquhar,
  Foerster, and Whiteson}{Rashid et~al\mbox{.}}{2018}]%
        {qmix}
\bibfield{author}{\bibinfo{person}{Tabish Rashid}, \bibinfo{person}{Mikayel
  Samvelyan}, \bibinfo{person}{Christian~Schr{\"{o}}der de Witt},
  \bibinfo{person}{Gregory Farquhar}, \bibinfo{person}{Jakob~N. Foerster},
  {and} \bibinfo{person}{Shimon Whiteson}.} \bibinfo{year}{2018}\natexlab{}.
\newblock \showarticletitle{{QMIX:} Monotonic Value Function Factorisation for
  Deep Multi-Agent Reinforcement Learning}. In
  \bibinfo{booktitle}{\emph{Proceedings of ICML}}.
\newblock


\bibitem[\protect\citeauthoryear{Samvelyan, Rashid, de~Witt, Farquhar,
  Nardelli, Rudner, Hung, Torr, Foerster, and Whiteson}{Samvelyan
  et~al\mbox{.}}{2019}]%
        {smac}
\bibfield{author}{\bibinfo{person}{Mikayel Samvelyan}, \bibinfo{person}{Tabish
  Rashid}, \bibinfo{person}{Christian~Schr{\"{o}}der de Witt},
  \bibinfo{person}{Gregory Farquhar}, \bibinfo{person}{Nantas Nardelli},
  \bibinfo{person}{Tim G.~J. Rudner}, \bibinfo{person}{Chia{-}Man Hung},
  \bibinfo{person}{Philip H.~S. Torr}, \bibinfo{person}{Jakob~N. Foerster},
  {and} \bibinfo{person}{Shimon Whiteson}.} \bibinfo{year}{2019}\natexlab{}.
\newblock \showarticletitle{The StarCraft Multi-Agent Challenge}. In
  \bibinfo{booktitle}{\emph{Proceedings of AAMAS}}.
\newblock


\bibitem[\protect\citeauthoryear{Son, Kim, Kang, Hostallero, and Yi}{Son
  et~al\mbox{.}}{2019}]%
        {qtran}
\bibfield{author}{\bibinfo{person}{Kyunghwan Son}, \bibinfo{person}{Daewoo
  Kim}, \bibinfo{person}{Wan~Ju Kang}, \bibinfo{person}{David Hostallero},
  {and} \bibinfo{person}{Yung Yi}.} \bibinfo{year}{2019}\natexlab{}.
\newblock \showarticletitle{{QTRAN:} Learning to Factorize with Transformation
  for Cooperative Multi-Agent Reinforcement Learning}. In
  \bibinfo{booktitle}{\emph{Proceedings of ICML}}.
\newblock


\bibitem[\protect\citeauthoryear{Strehl and Littman}{Strehl and
  Littman}{2008}]%
        {mbie-eb}
\bibfield{author}{\bibinfo{person}{Alexander~L. Strehl} {and}
  \bibinfo{person}{Michael~L. Littman}.} \bibinfo{year}{2008}\natexlab{}.
\newblock \showarticletitle{An analysis of model-based Interval Estimation for
  Markov Decision Processes}.
\newblock \bibinfo{journal}{\emph{J. Comput. System Sci.}}
  \bibinfo{volume}{74}, \bibinfo{number}{8} (\bibinfo{year}{2008}),
  \bibinfo{pages}{1309 -- 1331}.
\newblock


\bibitem[\protect\citeauthoryear{Sunehag, Lever, Gruslys, Czarnecki, Zambaldi,
  Jaderberg, Lanctot, Sonnerat, Leibo, Tuyls, and Graepel}{Sunehag
  et~al\mbox{.}}{2018}]%
        {vdn}
\bibfield{author}{\bibinfo{person}{Peter Sunehag}, \bibinfo{person}{Guy Lever},
  \bibinfo{person}{Audrunas Gruslys}, \bibinfo{person}{Wojciech~Marian
  Czarnecki}, \bibinfo{person}{Vin{\'{\i}}cius~Flores Zambaldi},
  \bibinfo{person}{Max Jaderberg}, \bibinfo{person}{Marc Lanctot},
  \bibinfo{person}{Nicolas Sonnerat}, \bibinfo{person}{Joel~Z. Leibo},
  \bibinfo{person}{Karl Tuyls}, {and} \bibinfo{person}{Thore Graepel}.}
  \bibinfo{year}{2018}\natexlab{}.
\newblock \showarticletitle{Value-Decomposition Networks For Cooperative
  Multi-Agent Learning}. In \bibinfo{booktitle}{\emph{Proceedings of AAMAS}}.
\newblock


\bibitem[\protect\citeauthoryear{Tan}{Tan}{1993}]%
        {iql}
\bibfield{author}{\bibinfo{person}{Ming Tan}.} \bibinfo{year}{1993}\natexlab{}.
\newblock \showarticletitle{Multi-Agent Reinforcement Learning: Independent vs.
  Cooperative Agents}. In \bibinfo{booktitle}{\emph{Proceedings of ICML}}.
\newblock


\bibitem[\protect\citeauthoryear{Tang, Houthooft, Foote, Stooke, Chen, Duan,
  Schulman, Turck, and Abbeel}{Tang et~al\mbox{.}}{2017}]%
        {hashtag_exploration}
\bibfield{author}{\bibinfo{person}{Haoran Tang}, \bibinfo{person}{Rein
  Houthooft}, \bibinfo{person}{Davis Foote}, \bibinfo{person}{Adam Stooke},
  \bibinfo{person}{Xi Chen}, \bibinfo{person}{Yan Duan}, \bibinfo{person}{John
  Schulman}, \bibinfo{person}{Filip~De Turck}, {and} \bibinfo{person}{Pieter
  Abbeel}.} \bibinfo{year}{2017}\natexlab{}.
\newblock \showarticletitle{{\#}Exploration: {A} Study of Count-Based
  Exploration for Deep Reinforcement Learning}. In
  \bibinfo{booktitle}{\emph{Proceedings of NeurIPS}}.
\newblock


\bibitem[\protect\citeauthoryear{Wang, Wang, Wu, and Zhang}{Wang
  et~al\mbox{.}}{2020}]%
        {edti}
\bibfield{author}{\bibinfo{person}{Tonghan Wang}, \bibinfo{person}{Jianhao
  Wang}, \bibinfo{person}{Yi Wu}, {and} \bibinfo{person}{Chongjie Zhang}.}
  \bibinfo{year}{2020}\natexlab{}.
\newblock \showarticletitle{Influence-Based Multi-Agent Exploration}. In
  \bibinfo{booktitle}{\emph{Proceedings of ICLR}}.
\newblock


\end{thebibliography}


\end{document}